\documentclass[preprint,review,12pt]{elsarticle}
\graphicspath{ {./}{./figures/} }
\usepackage{hyperref}
\usepackage{verbatim}
\usepackage{amsmath}
\usepackage{booktabs}
\usepackage{subcaption}
\usepackage{amssymb}
\journal{Digital Signal Processing}

\begin{document}
\begin{frontmatter}

\title{SMDDFNet: State-space Modeling and Dynamic Dual Fusion Network for Traffic Sign Detection}

\author[label1,label2]{Tianyi Yu}
\ead{tianyiyu@zju.edu.cn}
\author[label2]{Dajian Zhong \corref{cor1}}
\ead{djzhong@shmtu.edu.cn}
\author[label1]{Lilin Wang \corref{cor1}}
\ead{lilin.wang@zju.edu.cn}

\cortext[cor1]{Corresponding authors. Primary contact: Dajian Zhong. Tel.: +86-133-8619-7079.}
\address[label1]{State Key Laboratory of Ocean Sensing \& Ocean College, Zhejiang University, Zhoushan, 316021, China}
\address[label2]{College of Information Engineering, Shanghai Maritime University, Shanghai, 201306, China}

\begin{abstract}
Traffic sign detection is a challenging visual signal processing task for advanced driver assistance, where small objects, scale variation, and occlusion limit conventional detectors with fixed receptive fields. This paper proposes State-space Modeling and Dynamic Dual Fusion Network (SMDDFNet), a deep learning detector for traffic sign images. SMDDFNet integrates a Dynamic Dual Fusion (DDF) module and a state-space modeling backbone to enhance multi-scale feature representation. DDF combines efficient multi-scale attention with content-aware dynamic filtering in the frequency domain, while the backbone captures long-range dependencies with linear computational complexity. A multi-scale feature fusion neck further aggregates pyramid features for robust localization of small signs. Experiments on TT100K, GTSDB, PASCAL VOC, and the Roboflow~100 \emph{vehicle} subset show that SMDDFNet achieves competitive accuracy against recent detectors while retaining real-time throughput. The source code is available at \url{https://github.com/rainbowyuyu/SMDDFNet}.
\end{abstract}


\end{frontmatter}

\section{Introduction}
\begin{figure}[!t]
    \centering
    \includegraphics[width=\linewidth,keepaspectratio]{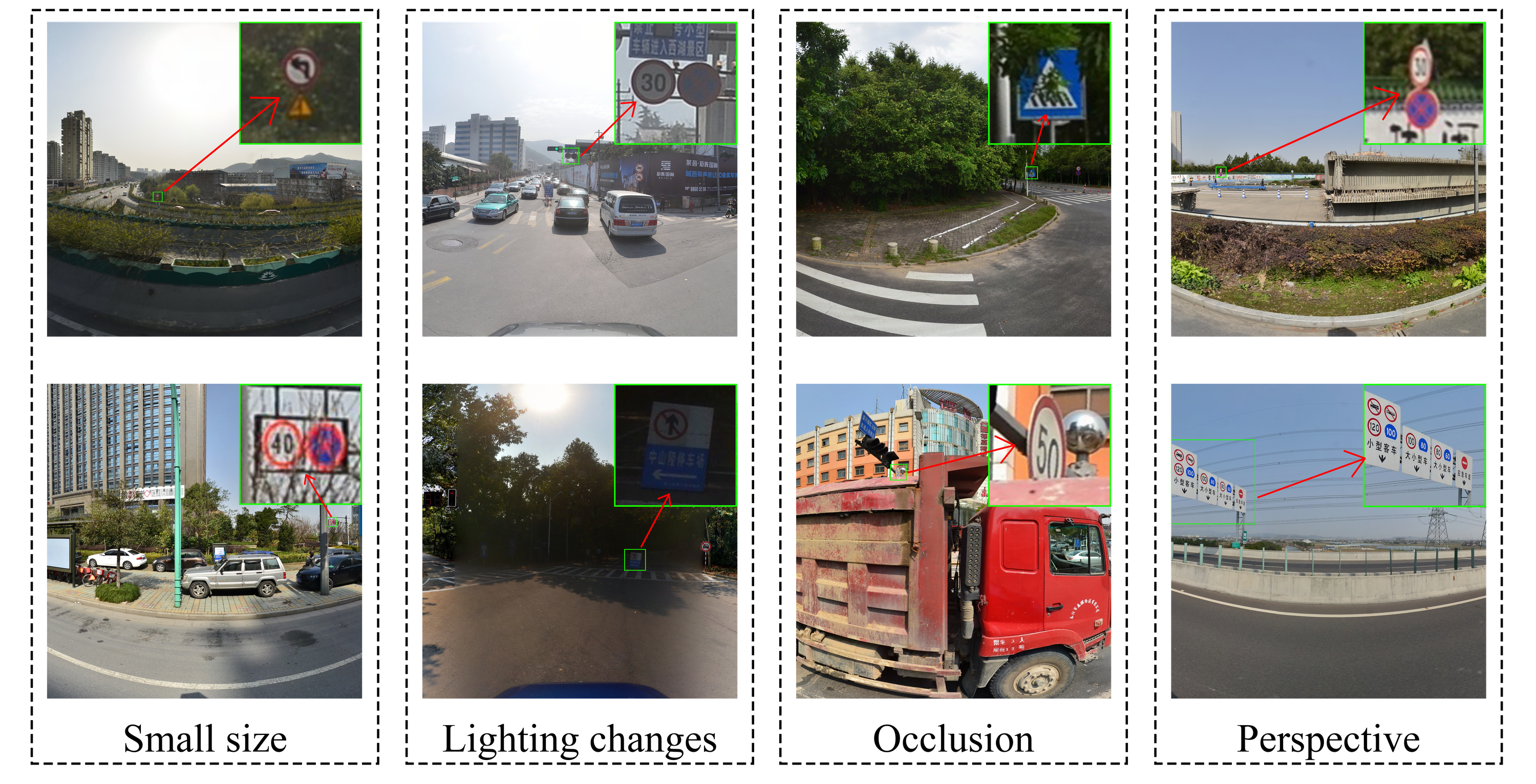}
    \caption{Significant challenges in detecting traffic signs including small size, lighting changes, occlusions, and different viewing angles.}
    \label{fig:likes}
\end{figure}

Traffic sign detection (TSD) is an important visual signal processing problem for advanced driver-assistance systems (ADAS) and autonomous driving, where detectors must remain accurate under small object scale, intra-class variation, and adverse imaging conditions. Traffic signs convey essential regulatory, warning, and guidance information. The real-time and accurate recognition of traffic signs is vital for ensuring road safety and enabling intelligent decision-making in autonomous vehicles~\cite{stallkamp2012man}. Despite significant progress in deep learning-based object detectors, TSD remains a challenging problem due to several real-world factors~\cite{Zhe_2016_CVPR}. Specifically, traffic signs often appear as small-scale objects within images, exhibit significant intra-class variation in shape, color, and style, and are frequently affected by challenging conditions such as illumination changes, partial occlusion, motion blur, and varying viewpoints, as illustrated in Figure~\ref{fig:likes}.

Conventional traffic sign detection (TSD) methods predominantly rely on convolutional neural networks (CNNs) with fixed receptive fields and manually designed feature fusion strategies. For instance, Faster R-CNN \cite{ren2015faster} employs a region proposal network with a standard backbone to generate object proposals. R-FCN \cite{dai2016r} uses position-sensitive score maps to improve localization. SSD \cite{liu2016ssd} and YOLOv2 \cite{redmon2016you} adopt single-stage multi-scale detection pipelines to accelerate inference. Some methods, such as RefineDet \cite{zhang2018single} and RON \cite{kong2017ron}, introduce cascaded refinement or hierarchical feature aggregation to enhance small object detection. Despite achieving promising results in general object detection, these approaches exhibit two main limitations when applied to TSD. First, standard backbone networks often struggle to extract sufficiently diverse and discriminative features for small, distant, or partially occluded traffic signs. This limitation arises because conventional convolutional layers have a restricted receptive field and limited ability to model long-range dependencies, which hampers the capture of rich contextual information. Second, existing fusion mechanisms are typically static and hand-crafted, which makes them inadequate for handling large scale variations inherent in traffic sign datasets. Fixed feature fusion strategies may fail to fully exploit multi-resolution information, resulting in suboptimal performance when detecting small or faraway signs. Consequently, while effective for general object detection, these conventional CNN-based methods are less suited for the unique challenges presented by TSD, motivating the development of adaptive multi-scale and context-aware fusion techniques.

To deal with the above challenges, this work is motivated by two key observations: (1) the need to capture long-range dependencies and global contextual information in a lightweight manner, and (2) the importance of dynamically enhancing multi-scale features to better represent fine-grained semantic details. Therefore, we propose SMDDFNet, a model specifically designed to overcome the unique difficulties of TSD. By incorporating a State-space Modeling Network, SMDDFNet efficiently handles long-range contextual information with linear complexity, addressing common issues such as varying lighting conditions and changing viewpoints, which are crucial for accurate detection in dynamic environments.

The core motivation behind SMDDFNet is to bridge the gap between high-accuracy detection and real-time performance, which is especially important in autonomous driving systems. For autonomous vehicles, traffic sign detection must not only be accurate but also fast, as real-time decision-making is essential for safe driving. SMDDFNet provides a practical solution for real-time TSD by improving detection accuracy while maintaining high computational efficiency, making it ideal for deployment in intelligent traffic systems and autonomous vehicles. The contributions of this work are summarized as follows:

\begin{itemize}
\item \textbf{We propose a lightweight yet high performance network named SMDDFNet to detect small traffic signs.} We introduce a state-space modeling backbone that captures long-range dependencies with linear complexity. Compared to traditional CNN or Transformer-based architectures, state-space modeling achieves better contextual feature encoding at a significantly lower computational cost, making this model suitable for real-time deployment in resource-constrained systems.
    
\item \textbf{A novel Dynamic Dual Fusion (DDF) module is designed to enhance the feature representation.} The DDF module integrates two complementary mechanisms: Efficient Multi-scale Attention (EMA) and content-aware Dynamic Filtering. These components are organized in a dual-branch architecture, allowing the model to simultaneously enhance spatial resolution and semantic richness. The adaptive fusion process enables robust detection of small and occluded objects under diverse scenes.
    
\item \textbf{Multi-scale feature learning is enhanced through efficient and adaptive fusion mechanisms.} By leveraging both fine-grained details from low-resolution features and high-level semantics, SMDDFNet ensures effective integration of multi-scale information. This approach minimizes the risk of missing small or distant signs, while improving the ability of the detector to generalize across diverse environments and varying object scales.

\item \textbf{The effectiveness of the proposed method is validated through extensive experiments.} Both qualitative and quantitative evaluations demonstrate that SMDDFNet achieves superior detection accuracy compared to state-of-the-art methods while maintaining real-time inference speed. Experimental results confirm the effectiveness of the backbone and fusion strategy in addressing the unique challenges of traffic sign detection.
\end{itemize}

\section{Related work}

\subsection{Backbone networks}
Backbone networks play a fundamental role in object detectors by extracting hierarchical and semantically rich representations. Early works such as VGGNet~\cite{simonyan2014very} and ResNet~\cite{he2016deep} demonstrated strong feature extraction capabilities but incurred high computational costs due to their deep and heavy architectures. To improve efficiency, lightweight models like MobileNet~\cite{howard2017mobilenets}, ShuffleNet~\cite{zhang2018shufflenet}, and GhostNet~\cite{han2020ghostnet} were proposed, offering a balance between speed and accuracy. More recently, attention-based models such as DETR~\cite{carion2020end} and Swin Transformer~\cite{liu2021swin} achieved competitive performance via long-range dependency modeling, although their high resource consumption limits deployment in real-time scenarios. To overcome these limitations, the SSM architecture~\cite{gu2023mamba} introduces a state-space model with linear-time sequence modeling capability, enabling efficient extraction of both local and global features. Zaidi et al.~\cite{zaidi2022dsp_survey} surveyed modern deep learning object detectors and summarized backbone designs for real-time recognition. Qiu et al.~\cite{qiu2025dpyolo} further proposed DP-YOLO, a lightweight traffic sign detector tailored to small objects in complex driving scenes.

\subsection{Multi-Scale feature fusion}
Accurate detection of small-scale traffic signs remains a major challenge due to their limited pixel footprint and the loss of spatial resolution in deep layers. Multi-scale feature fusion techniques have been extensively explored to enhance the detection of small-scale signs, which often suffer from limited resolution and spatial degradation. Classical methods such as FPN~\cite{lin2017feature} and PANet~\cite{liu2018path} fuse features across scales, while more advanced methods like NAS-FPN~\cite{ghiasi2019fpn} and BiFPN~\cite{tan2020efficientdet} optimize this process using architecture search and bidirectional pathways. Despite their effectiveness, these approaches often introduce significant computation. Han et al.~\cite{han2024ednyolo} proposed EDN-YOLO with an efficient decoupled head and NWD-enhanced localization loss for multi-scale traffic sign detection on TT100K and GTSDB. Chen et al.~\cite{chen2024cspgnet} introduced CSPGNet for cross-scale spatial perception of tiny objects, providing a complementary perspective on multi-scale fusion for small-target detection.

\subsection{Dynamic feature refinement techniques}
Dynamic feature refinement has gained attention for its ability to adaptively modulate feature maps based on contextual cues. Representative approaches include SE-Net~\cite{hu2018squeeze}, CBAM~\cite{woo2018cbam}, and Deformable DETR~\cite{zhu2020deformable}, which recalibrate features using attention or deformable mechanisms. Dynamic convolutions~\cite{chen2020dynamic} and DCNs~\cite{dai2017deformable} provide content-aware transformations to enhance feature learning. Jing et al.~\cite{jing2024dynamic} proposed a Dynamic Feature Focusing Network (DFFN) specifically designed for small object detection. Their approach introduces modules such as the Visual Perception Enhancement Module (VPEM) and a Task Association module, which dynamically emphasize informative features while suppressing irrelevant background noise. This design effectively improves the detection of small-scale objects under complex and challenging conditions. In contrast to existing works, the proposed method introduces a lightweight yet effective Dynamic Dual Fusion (DDF) module that integrates spatial and semantic information using efficient multi-kernel convolutions and attention-based filtering. By incorporating Efficient Multi-scale Attention (EMA)~\cite{ouyang2023efficient} and Dynamic Filter (DF)~\cite{chen2020dynamic} mechanisms in a dual-path design, this work enables adaptive feature modulation and accurate small object detection under various challenging conditions, while maintaining real-time performance. The key difference from prior works lies in the unified design of DDF, which simultaneously addresses scale variation and context adaptation through lightweight, efficient modules, making it well-suited for real-world TSD scenarios.

\section{Proposed method}
In this paper, we propose SMDDFNet, a novel end-to-end detection framework that integrates a State-space Modeling Network with a Dynamic Dual Fusion (DDF) module and a Multi-Scale Feature Fusion (MSFF) module to address these challenges. 

\subsection{Network overview}
\begin{figure}[!t]
\centering
\includegraphics[width=1\textwidth]{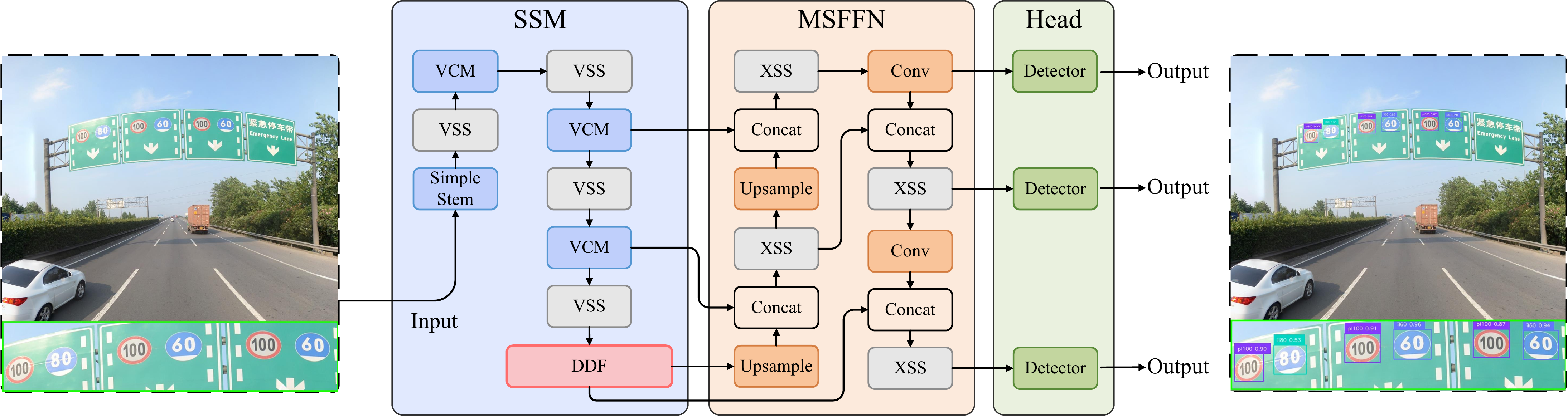}
\caption{The architecture of SMDDFNet, comprising a state-space modeling (SSM) network, a Multi-Scale Feature Fusion (MSFF) module, and detection heads (Head). In the SSM network, the proposed DDF module combines the EMA and DF modules for adaptive feature enhancement.}
\label{fig:all_net}
\end{figure}

As illustrated in Figure~\ref{fig:all_net}, the proposed SMDDFNet is an end-to-end architecture tailored for traffic sign detection, emphasizing efficient feature extraction and dynamic multi-scale representation. Unlike the conventional YOLOv8 backbone, SMDDFNet employs a State-space Modeling Network leveraging Vision State Space Blocks (VSSBlocks) to efficiently model spatial dependencies, enabling the model to operate with linear complexity. This ensures high computational efficiency, making it particularly suitable for real-time applications where speed is critical.

To further enhance the representation of high-level features, the original SPPF module is replaced with the novel Dynamic Dual Fusion (DDF) module, composed of Efficient Multi-scale Attention (EMA) and Dynamic Filter (DF). EMA extracts and aggregates hierarchical context features through grouped convolutions and attention-based gating mechanisms, while DF dynamically adjusts convolutional kernels based on input content, enabling adaptive and fine-grained feature modulation. These capabilities significantly improve detection accuracy, especially for small object sizes and occlusions.

The extracted multi-scale features are passed through the Multi-Scale Feature Fusion (MSFF) module, where feature fusion and refinement are conducted. The MSFF integrates features at different scales using Cross Scale Summation Blocks (XSS), effectively merging fine-grained spatial detail from shallow stages with high-level semantics from deeper stages. This fusion ensures robust integration of multi-scale information, minimizing the risk of missing small or distant signs and enhancing the model's ability to generalize across diverse environments and varying object scales. Finally, multiple detection heads perform classification and localization tasks. By combining the strengths of state-space modeling and DDF, SMDDFNet achieves more accurate and robust traffic sign detection, particularly in challenging real-world conditions such as varying object scales, lighting, and occlusions.

\subsection{State-space modeling network}
To enhance representation capacity and long-range dependency modeling in traffic sign detection, the proposed method adopts a State-space Modeling Network as the backbone of SMDDFNet. Inspired by the recently proposed SSM architecture~\cite{gu2023mamba}, which efficiently models long-sequence dependencies via selective spatial state-space modeling, the backbone leverages Vision State Space Blocks (VSSBlocks) for spatial-aware feature extraction with linear computational complexity.

The backbone architecture is composed of the following key components. SimpleStem is a lightweight convolutional stem used to project input images into a lower-resolution, high-dimensional feature space, enhancing early-stage representations. VSSBlocks serve as the fundamental units in each stage. They integrate the selective spatial mixing mechanism of state-space modeling to capture both local and global dependencies without attention, improving model efficiency. Vision Clue Merge is a transition module inserted between stages to downsample the spatial resolution and double the channel dimension while fusing adjacent stage features. Unlike traditional backbones that end with pooling layers such as SPP or SPPF, the proposed method introduces the DDF module at the final stage. DDF dynamically fuses multi-kernel convolutions and dual-branch pathways to enhance feature richness and semantic completeness. It enables the model to better handle the scale and appearance variations commonly seen in traffic sign detection. The final representation from the backbone can be formulated as:

\begin{equation}
F_{\text{out}} = \text{DDF}\left(\text{VSS}_{L}\!\left(\cdots \text{VSS}_{1}\!\left(\text{Stem}(X)\right)\cdots\right)\right)
\end{equation}

where $\text{Stem}(X)$ denotes the initial embedding of the input image $X$, each $\text{VSS}_{i}$ is a Vision State Space Block, $\text{VSS}_{L}$ denotes the stacked blocks at the deepest stage, and DDF denotes the final fusion module. This State-space Modeling Network ensures high efficiency and robustness in extracting diverse and informative features, especially when applied to small object detection tasks such as traffic sign recognition.

\subsection{Dynamic dual fusion module}
\begin{figure}[!t]
\centering
\includegraphics[width=1\textwidth]{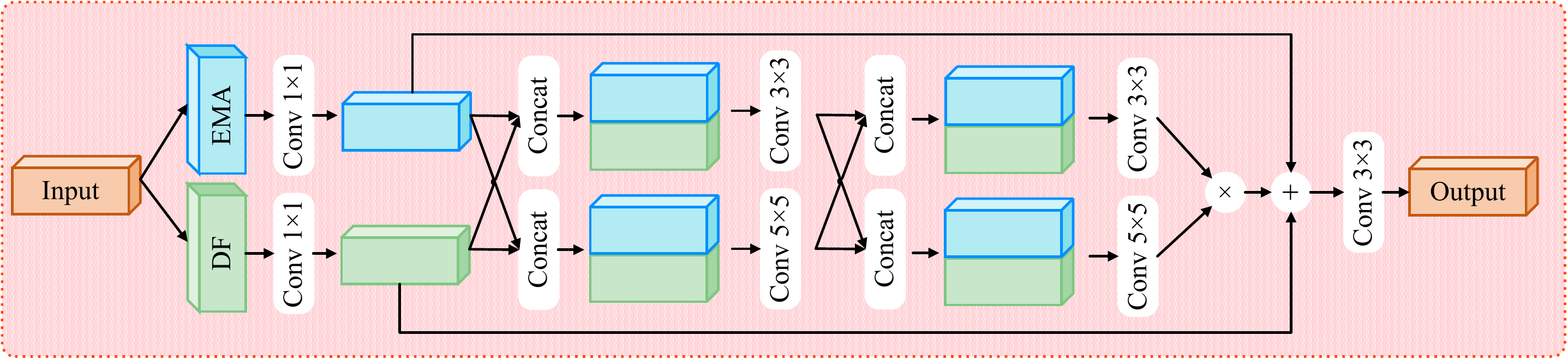}
\caption{Dynamic dual fusion (DDF) module for adaptive feature enhancement, integrating efficient multi-scale attention (EMA) and dynamic filter (DF).}
\label{fig:ddf}
\end{figure}

As illustrated in Figure~\ref{fig:ddf}, this paper proposes integrating a Dynamic Dual Fusion (DDF) module into SMDDFNet so that the backbone can effectively fuse multi-scale contextual features and improve feature discriminability. The DDF module adopts a dual-path fusion strategy, dynamically combining fine-grained and coarse-grained information through hierarchical convolutions, thereby capturing richer spatial and semantic features. The DDF module consists of two key components: Efficient Multi-scale Attention (EMA) and Dynamic Filter (DF), which are responsible for modeling spatial-directional attention and frequency-domain dynamic responses, respectively. These are ultimately concatenated and fused to generate an enhanced feature representation:

\begin{equation}
\text{DDF}(F) = \phi\left(\text{Con}\left(\psi_{3\times3}(F_{\text{EMA}}), \psi_{5\times5}(F_{\text{EMA}}), \psi_{3\times3}(F_{\text{DF}}), \psi_{5\times5}(F_{\text{DF}})\right)\right)
\end{equation}

\subsubsection{Efficient multi-scale attention}

\begin{figure}[!t]
\centering
\includegraphics[width=0.7\textwidth]{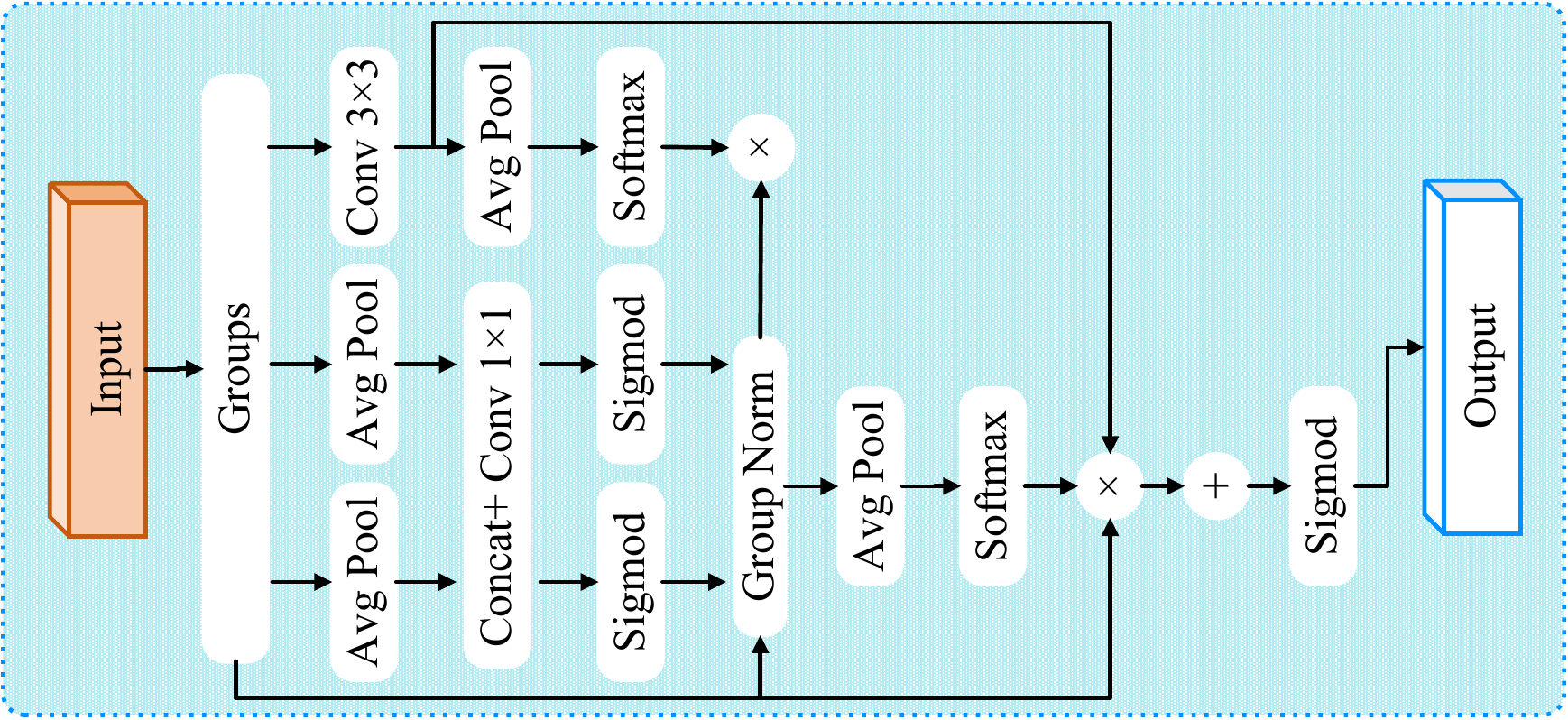}
\caption{Architecture of the EMA module for spatial structure-aware attention within the DDF framework.}
\label{fig:ema}
\end{figure}

As illustrated in the left part of Figure~\ref{fig:ema}, the core idea of EMA is to introduce a group-channel partitioning strategy and adaptive modeling in spatial directions. Specifically, the input feature map is first divided into several groups along the channel dimension. Each group is processed independently to reduce computational complexity and improve representational capacity. For each group, EMA performs adaptive average pooling along the horizontal and vertical directions to extract global spatial structural information. These direction-specific features are then fused via 1×1 convolution and passed through a Sigmoid activation to generate spatial attention weights, which are used to modulate the original feature map. Additionally, a parallel 3×3 convolution branch is introduced to enhance spatial features, which are then interactively fused with the attention-weighted features. Finally, EMA extracts a global context vector via global average pooling to compute guiding weights between the two branches, which are applied to the original features through element-wise multiplication, thus achieving a spatial structure-aware attention mechanism:

\begin{equation}
F_{\text{EMA}} = W \cdot \text{GN}(F_g \cdot A) + (1 - W) \cdot \text{Conv}_{3\times3}(F_g)
\end{equation}

where $W = \sigma(w_1 \cdot F_{\text{local}} + w_2 \cdot F_{\text{att}})$, the weights are computed after global average pooling (GAP), and $A$ is the EMA spatial attention map.

\subsubsection{Dynamic filter}

\begin{figure}[!t]
\centering
\includegraphics[width=0.6\textwidth]{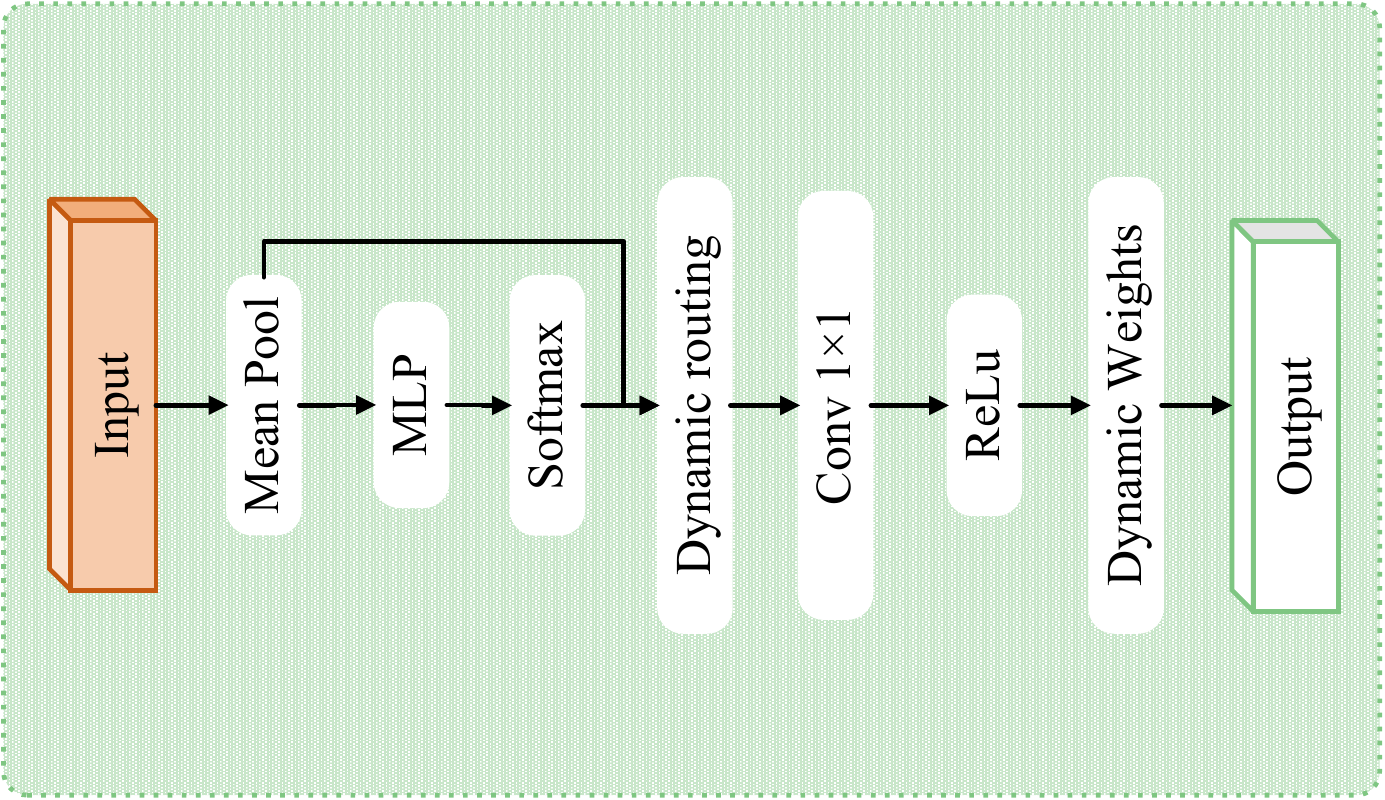}
\caption{Architecture of the DF module for frequency-domain response adaptation within the DDF framework.}
\label{fig:df}
\end{figure}

As illustrated in Figure~\ref{fig:df}, the Dynamic Filter (DF) module models the input feature map response variations from a frequency-domain perspective, adaptively generating convolution kernels via learned complex-valued dynamic weights for different scenes. First, the input spatial features are mapped to a higher-dimensional intermediate space via two pointwise (1×1) convolution layers and enhanced by the StarReLU activation function to increase non-linear expressiveness. Then, the feature map is transformed to the frequency domain using a 2D real Fourier transform (rFFT), where it undergoes element-wise multiplication with learnable complex dynamic filter kernels. These kernels are selected dynamically via a routing mechanism, where a router computes weights for each filter based on the global average response of the input, enabling input-adaptive selection. Finally, the filtered frequency-domain features are transformed back to the spatial domain using the inverse Fourier transform (iFFT), and the original channel structure is restored. This process effectively generates a dedicated set of filters for each input feature map, enhancing the responsiveness of the model to varying patterns, textures, and shapes:

\begin{equation}
F_{\text{DF}} = \phi_2\left(\text{iFFT}\left(\text{rFFT}\left(\phi_1(F)\right) \cdot \sum_{i=1}^{N} r_i \cdot W_i\right)\right)
\end{equation}

where $W_i$ are the frequency-domain filter kernels, $r_i$ are the dynamic weights generated by the router, and $\phi_1, \phi_2$ are the input and output mapping functions, respectively.

The DDF module first splits the input feature map into two branches, which are processed by the EMA and DF modules to produce $F_{\text{EMA}}$ and $F_{\text{DF}}$, respectively. These are then compressed via 1×1 convolutions and concatenated, followed by 3×3 and 5×5 convolutions to extract multi-scale contextual information and perform cross-scale integration. Finally, the fused feature map is combined with the individual outputs of EMA and DF branches via weighted addition, and passed through a 3×3 convolution to produce the final enhanced representation:

\begin{equation}
\text{Output} = \text{Conv}_{3\times3}\left(F_{\text{fused}} + \alpha \cdot F_{\text{EMA}} + \beta \cdot F_{\text{DF}}\right)
\end{equation}

where $F_{\text{fused}}$ is the output of the multi-scale convolution fusion, and $\alpha$, $\beta$ are learnable fusion weight coefficients.

By integrating multi-scale features from two independent transformation branches, the Dynamic Dual Fusion module captures richer spatial and contextual information, enabling the network to better recognize objects under complex visual conditions. Each branch processes the input feature map through distinct convolutional operations, extracting information from different receptive fields. This dual-path architecture not only increases the diversity of learned features but also facilitates the integration of complementary cues such as edges, textures, and high-level semantics.

\subsection{Multi-scale feature fusion module}

The Multi-Scale Feature Fusion (MSFF) module is designed to enhance feature extraction across different spatial resolutions by progressively merging multi-scale features from the backbone. The process improves the ability of the network to capture both fine details and high-level semantics, particularly useful for tasks like traffic sign detection. The MSFF consists of several stages, where the input image passes through layers that progressively reduce the spatial resolution while refining the feature maps. At each stage, the resolution changes as follows: at stage \( P1 \), the resolution is \( H \times W \), at \( P2 \) it becomes \( \frac{H \times W}{4} \), at \( P3 \) it reduces to \( \frac{H \times W}{8} \), at \( P4 \) it further decreases to \( \frac{H \times W}{16} \), and at \( P5 \), the resolution reaches \( \frac{H \times W}{32} \). This multi-scale process enables the network to capture both fine-grained details and global context, significantly enhancing its detection capabilities.

\subsubsection{Upsampling and fusion}
The first step in the MSFF involves upsampling the feature map from a lower resolution to match the resolution of the higher-level features. This is done using nearest-neighbor upsampling, which efficiently increases the spatial resolution without introducing unnecessary computational complexity. The upsampled feature map is then concatenated with the feature map from the corresponding stage in the backbone, ensuring that both fine-grained details and global context are retained for further processing:

\begin{equation}
\text{F}_{\text{upsampled}} = \text{Upsample}(\text{F}_{\text{low-res}}, \text{scale}=2, \text{mode}='nearest')
\end{equation}

\subsubsection{Feature concatenation}
After the upsampling step, the resulting feature map is concatenated with the higher-level feature map from the backbone. This concatenation merges information from different levels of the network, allowing the model to benefit from both the fine details of low-resolution features and the abstract semantic features from high-resolution maps. The concatenated feature map is then passed through a Cross Scale Summation Block (XSSBlock) for spatial-aware feature extraction, enabling the model to refine both local and global dependencies:

\begin{equation}
\text{F}_{\text{fused}} = \text{Concat}(\text{F}_{\text{upsampled}}, \text{F}_{\text{high-res}}, \text{dim}=1)
\end{equation}

\subsubsection{Multi-scale feature refinement}
The XSSBlock processes the concatenated features by applying a multi-scale spatial feature extraction technique. It is designed to adapt to various object sizes by using different channel dimensions for each scale. The process ensures that the network can capture fine details at smaller scales (e.g., small traffic signs) and retain global context at larger scales (e.g., background information).
For instance, for smaller objects, the network uses a reduced number of channels and higher resolution to focus on fine details:

\begin{equation}
\text{F}_{\text{small}} = \text{XSSBlock}(\text{F}_{\text{P3}}, 256)
\end{equation}

For medium objects, the network uses a higher number of channels and a mid-level resolution to balance detail and context:

\begin{equation}
\text{F}_{\text{medium}} = \text{XSSBlock}(\text{F}_{\text{P4}}, 512)
\end{equation}

For large objects, the network employs a lower resolution and even more channels, focusing on global context:

\begin{equation}
\text{F}_{\text{large}} = \text{XSSBlock}(\text{F}_{\text{P5}}, 1024)
\end{equation}

\subsubsection{Adaptive feature fusion}
After each feature block processes the different object sizes, the resulting feature maps are fused at each level. This fusion process combines the rich, high-level semantics from the large objects with the finer details of small objects. The fusion ensures that the model maintains spatial consistency and the correct feature representations across multiple scales. The output from each stage is prepared for the final detection task.

\subsection{Detection heads}
Following the MSFF module, SMDDFNet attaches multiple detection heads at pyramid levels \(P_3\), \(P_4\), and \(P_5\), following a YOLO-style single-stage layout. Each head uses separate branches for bounding-box regression and class scoring over traffic-sign categories, taking fused multi-scale features from the neck so that small or distant signs are handled at higher spatial resolution while larger instances exploit deeper semantics.

\subsection{Feature map size transformation in SMDDFNet}
\begin{figure}[!t]
\centering
\includegraphics[width=1\textwidth]{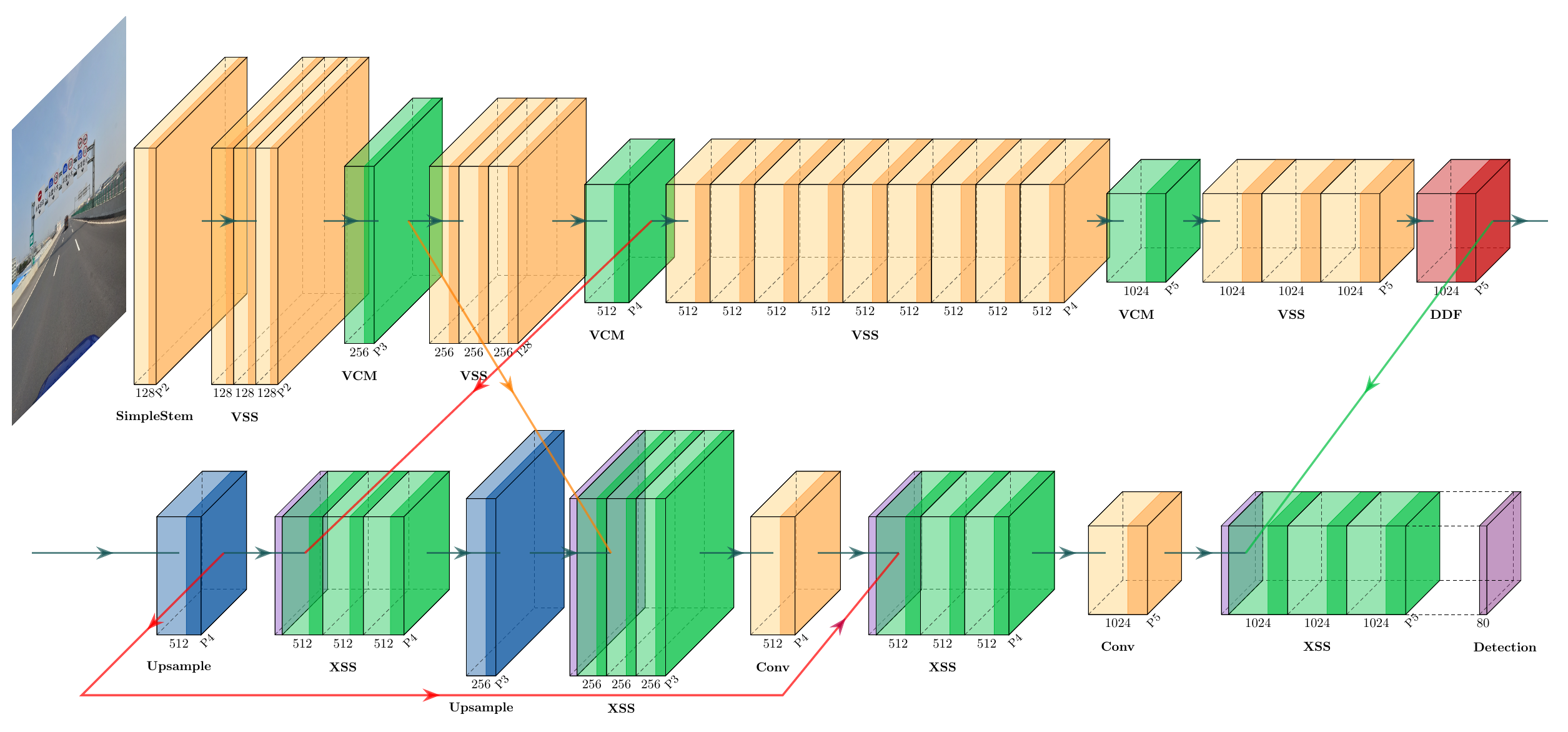}
\caption{Feature map size transformation in SMDDFNet.}
\label{fig:feature}
\end{figure}

The SMDDFNet architecture processes an input image of spatial size \(H \times W\) through the backbone and neck, progressively changing spatial resolution and channel width until detection heads produce bounding boxes and class scores, as summarized in Figure~\ref{fig:feature}. Denoting \(P_k\) as the feature map at pyramid level \(k\) aligned with common YOLO-style heads, the backbone first applies SimpleStem to obtain early features at stride~4 with 128 channels (\(\frac{H}{4}\!\times\!\frac{W}{4}\)). VisionClueMerge and VSSBlocks then yield deeper scales \(P_3\), \(P_4\), and \(P_5\) with spatial sizes \(\frac{H}{8}\!\times\!\frac{W}{8}\), \(\frac{H}{16}\!\times\!\frac{W}{16}\), and \(\frac{H}{32}\!\times\!\frac{W}{32}\), respectively, and channel dimensions increasing to 256, 512, and 1024. The DDF module refines the deepest representation before multi-scale detection. In the neck, features at \(P_5\) are upsampled and fused with \(P_4\), then similarly merged toward \(P_3\) via concatenation and XSSBlocks, recovering finer spatial detail for small instances while retaining semantics from deeper layers. Detection outputs are produced at \(\{P_3,P_4,P_5\}\), enabling scale-aware localization across the scene. This hierarchical design yields robust detection under large appearance and scale variation.

\section{Experiment}
\subsection{Datasets}

To evaluate the effectiveness of the proposed SMDDFNet, experiments were conducted on multiple standard datasets, including TT100K, GTSDB, PASCAL VOC~\cite{everingham2010pascal}, and the Roboflow~100 \emph{vehicle} subset~\cite{ciaglia2022rf100}. These benchmarks, with their distinct object taxonomies and imaging conditions, help in comprehensively assessing the performance of SMDDFNet in traffic-sign, general object, and vehicle-centric scenarios.

\begin{figure}[!b]
    \centering
    \includegraphics[width=1\linewidth]{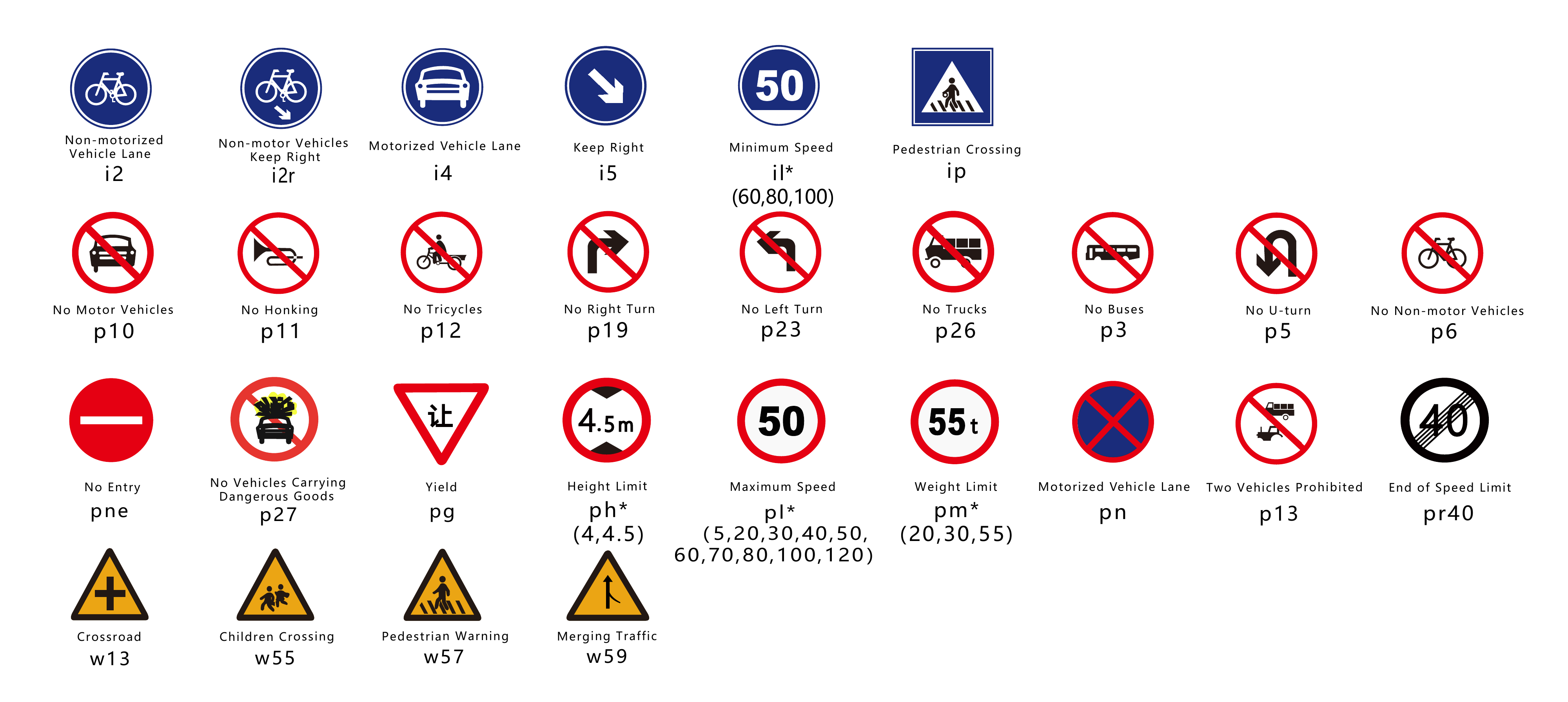}
    \caption{Diagram of the filtered subset of 42 traffic sign categories from the TT100K dataset, similar types of signs are grouped together for illustration.}
    \label{fig:enter-label}
\end{figure}

\subsubsection{TT100K dataset}
The TT100K dataset \cite{Zhe_2016_CVPR} is a widely-used benchmark for traffic sign detection, particularly in real-world driving scenarios. The dataset was collected from roadways in China using dashcams, covering a range of challenging environmental conditions such as low-light, motion blur, occlusions, and scale variations, all of which pose significant challenges for real-time detection systems. The TT100K dataset consists of over 10,000 images, spanning more than 200 traffic sign categories, with approximately 100 categories containing sufficient instances for effective training. Following common practice, we adopted the standard filtered subset, which includes 5,962 training images and 2,979 test images, focusing on 45 labels, as shown in Figure~\ref{fig:enter-label}. Each image has a resolution of 2048×2048, ensuring rich spatial detail while also increasing the computational burden, making it an ideal testbed for evaluating lightweight yet powerful backbones like SMDDFNet. Given the small object size and dense distribution of traffic signs in the TT100K dataset, models must be capable of precise multi-scale feature extraction and robust contextual reasoning. This aligns perfectly with the design goals of the State-space Modeling Network and the Dynamic Dual Fusion (DDF) module in SMDDFNet.

\subsubsection{GTSDB dataset}
The German Traffic Sign Detection Benchmark (GTSDB) is a single-image detection benchmark designed for researchers interested in computer vision, pattern recognition, and image-based driver assistance systems~\cite{houben2013gtsdb}. Introduced at the IEEE International Joint Conference on Neural Networks in 2013, GTSDB focuses on single-image traffic sign detection. The dataset consists of 900 images, divided into 600 training images and 300 evaluation images. It is divided into three categories, each catering to the specific properties of various detection approaches. These categories help accommodate models with different detection capabilities and properties. Additionally, an online evaluation system is available for immediate analysis and ranking of the submitted results. The images in the dataset contain zero to six traffic signs. Even if a traffic sign is present in an image, it may not belong to the competition-relevant categories, which include prohibitive, danger, and mandatory signs. The images are stored in PPM format, and the sizes of the traffic signs in the images range from 16×16 to 128×128 pixels. These signs may appear in different perspectives and under various lighting conditions.

\subsubsection{PASCAL VOC dataset}
\label{subsec:voc}
The PASCAL Visual Object Classes (VOC) challenge~\cite{everingham2010pascal} is a canonical detection benchmark covering twenty everyday object categories. Following common practice, we train on the union of VOC~2007 \texttt{trainval} and VOC~2012 \texttt{trainval} and evaluate on the VOC~2007 \texttt{test} split, reporting mean average precision at IoU $0.5$ (mAP@0.5) and at IoU $0.5{:}0.95$ (mAP@0.5:0.95). VOC provides a complementary view to traffic-sign datasets because it mixes scales, clutter, and non-sign backgrounds, testing whether the proposed fusion and backbone still generalize outside curated road scenes.

\subsubsection{Vehicle dataset}
\label{subsec:vehicle_dataset}
The \emph{vehicle} split belongs to the Roboflow~100 multi-domain detection benchmark~\cite{ciaglia2022rf100}. It is distributed together with download scripts and aggregated baseline scores through the official replication repository, while the corresponding imagery and hosted training exports remain available on Roboflow Universe. The split collects in-the-wild vehicle instances under diverse viewpoints, lighting, and partial occlusions. Following the benchmark protocol, we adopt the released train and validation partition, resize inputs to $640\!\times\!640$ to match the remainder of our pipeline, and report COCO-style mAP@0.5 and mAP@0.5:0.95 together with precision, recall, and F1 on the validation logs for methods where these diagnostics are available.

\subsection{Evaluation metrics}

To comprehensively evaluate the performance of the proposed SMDDFNet, this paper adopts three standard metrics: Average Precision (AP), computational cost in GFLOPs (giga floating-point operations per forward pass at the reported input resolution), and throughput in frames per second (FPS)~\cite{everingham2010pascal}.

Average Precision (AP) serves as the principal indicator of object detection accuracy. It measures the area under the precision-recall curve, reflecting the ability of the model to maintain high precision across varying recall levels. Formally, AP is defined as:
\begin{equation}
    \text{AP} = \int_{0}^{1} P(R) \, dR,
    \label{eq:ap}
\end{equation}
where $P(R)$ represents the precision as a function of recall $R$. To evaluate the overall detection performance across all object categories, this work computes Mean Average Precision (mAP)~\cite{lin2014microsoft}:
\begin{equation}
    \text{mAP@IoU} = \frac{1}{N} \sum_{i=1}^{N} \text{AP@IoU}_i,
    \label{eq:map}
\end{equation}
where $N$ is the total number of classes and $\text{IoU}_i$ denotes the Intersection over Union threshold for class $i$ and reports mAP@0.5, which measures detection performance at an IoU threshold of 0.5, as well as mAP@0.75 and mAP@[0.5:0.95] (averaged over IoU thresholds from 0.5 to 0.95 in 0.05 increments), providing a more comprehensive assessment across stricter criteria.

Additionally, this method is evaluated by $\text{AP}_S$, $\text{AP}_M$, and $\text{AP}_L$, which correspond to detection accuracy for small, medium, and large objects, respectively. GFLOPs quantify the computational cost of one forward pass at the stated input size. FPS measures inference throughput on our hardware, with higher values indicating faster processing. These metrics together offer a well-rounded evaluation of both the accuracy and efficiency of this network.

\subsection{Implementation details}
This paper implements SMDDFNet using PyTorch 2.6.0 and conducts all experiments on a single NVIDIA RTX 4090 GPU. To accelerate convergence, the model is initialized with pretrained weights from YOLOv8. The input images are resized to 640 × 640, and training is performed with a batch size of 8 for stable optimization. The model is trained for 300 epochs on the TT100K dataset, employing standard data augmentations such as Mosaic, MixUp, HSV jittering, and random affine transformations to improve generalization under real-world variations. The proposed method uses the Stochastic Gradient Descent (SGD) optimizer with an initial learning rate of 0.01, momentum of 0.937, and weight decay of $5 \times 10^{-4}$. A linear warm-up strategy is applied during the initial 3 epochs to stabilize training, followed by a cosine annealing schedule to gradually decay the learning rate. All experiments are conducted using float32 precision, and mixed precision training is optionally enabled to enhance speed and memory efficiency. PASCAL VOC and the \emph{vehicle} subset reuse the same optimizer, augmentation policy, and input resolution unless stated otherwise.

\subsection{Quantitative analysis}
We compare the proposed SMDDFNet with state-of-the-art traffic sign detection methods on the TT100K dataset. Table \ref{tab:tt100k_results} presents the performance of this model against various well-known detectors, evaluating them based on mAP@.5, mAP@.75, mAP@.5:.95, AP$_s$, AP$_m$, AP$_l$, FPS, and GFLOPs.

From Table~\ref{tab:tt100k_results}, SMDDFNet achieves 86.4\% mAP@.5, outperforming recent models such as YOLOv9 (86.5\%), Transformer Fusion (85.4\%), EMDFNet (84.4\%), Open-TransMind (83.4\%), and the YOLOv5s baseline trained on the augmented split reported by Lai et al.~\cite{lai2023stc} (79.6\%). It also attains 79.5\% mAP@.75, surpassing Cascade R-CNN (70.4\%), Faster R-CNN (69.3\%), and TRD-YOLO (67.2\%). Furthermore, SMDDFNet reaches 65.7\% mAP@.5:.95, highlighting its robustness across different IoU thresholds. A notable advantage of SMDDFNet is its capability in detecting small and medium-sized traffic signs. It achieves an AP$_s$ of 58.1\%, substantially higher than Sparse R-CNN (35.6\%), Faster R-CNN (27.7\%), and Cascade R-CNN (36.2\%). For medium objects, it achieves an AP$_m$ of 75.0\%, significantly surpassing Transformer Fusion (43.7\%). Regarding large objects, SMDDFNet maintains strong performance with an AP$_l$ of 83.5\%, outperforming Sparse R-CNN (70.6\%) and TRD-YOLO (40.2\%) among listed methods. In addition to accuracy, SMDDFNet demonstrates a favorable balance between speed and complexity. At $640\!\times\!640$ inputs, YOLOv5 attains a higher throughput (82.6 FPS) and VATSD reaches 100 FPS at $608\!\times\!608$, whereas SMDDFNet trades some FPS for substantially higher mAP@.5; it still runs faster than EMDFNet (46.7 FPS) and heavy two-stage detectors such as Cascade R-CNN (17.0 FPS) and Faster R-CNN (21.0 FPS). With 81.9 GFLOPs, SMDDFNet remains much more efficient than Sparse R-CNN (153.3 GFLOPs) and still competitive against Cascade R-CNN (239.2 GFLOPs). This indicates that the improved performance of SMDDFNet is achieved without excessive computational overhead, making it suitable for practical deployment. Table \ref{tab:tt100k_each_results} reports detection accuracy across different traffic sign categories, which further validate the effectiveness of the proposed method. SMDDFNet posts the strongest AP among listed methods on several challenging classes (e.g., \texttt{il60} at 98.3\%, \texttt{pne} at 95.8\%, and \texttt{il80} at 95.7\%), while remaining competitive on the remaining categories summarized in the table.

The effectiveness of SMDDFNet is attributed to enhanced feature extraction through an advanced backbone, a multi-scale fusion module that improves small object detection, and an efficient lightweight design that achieves superior accuracy while maintaining real-time performance.

\begin{table*}[t]
\centering
\caption{Comparison with existing methods on the TT100K dataset. }
\resizebox{\textwidth}{!}{
\begin{tabular}{lccccccccc}
\hline
\textbf{Model} & \textbf{Input Size} & \textbf{mAP@.5} & \textbf{mAP@.75} & \textbf{mAP@.5:.95} & \textbf{AP$_s$} & \textbf{AP$_m$} & \textbf{AP$_l$} & \textbf{FPS} & \textbf{GFLOPs} \\
\hline
Faster R-CNN \cite{ren2015faster} & 1024×1024 & 74.1 & 69.3 & 58.9 & 27.7 & 73.1 & 81.1 & 21 & 211.5 \\
SSD512 \cite{liu2016ssd} & 512×512 & 68.7 & - & - & - & - & - & - & - \\
Zhu et al.~\cite{Zhe_2016_CVPR} & 640×640 & 83.3 & \textit{79.0} & \textbf{65.8} & - & - & - & - & - \\
RetinaNet \cite{lin2017focal} & 2048×2048 & 61.9 & - & - & - & - & - & - & - \\
Mask R-CNN \cite{he2017mask} & 1000×800 & 70.8 & - & - & - & - & - & - & - \\
Cascade R-CNN \cite{cai2018cascade} & 1024×1024 & 81.2 & 70.4 & 63.5 & 36.2 & \textit{74.6} & \textit{83.0} & 17.0 & 239.2 \\
ScratchDet \cite{zhu2019scratchdet} & 512×512 & 74.0 & - & - & - & - & - & - & - \\
FCOS \cite{tian2019fcos} & 2048×2048 & 83.3 & - & - & - & - & - & - & - \\
CAB Net \cite{cui2020cabnet} & 512×512 & 78.0 & - & - & - & - & - & - & - \\
Sparse R-CNN \cite{sun2021sparse} & 1024×1024 & 65.0 & 62.8 & 53.2 & 35.6 & 61.2 & 70.6 & 21.0 & 153.3 \\
TRD-YOLO \cite{chu2023trd} & 512×512 & 73.5 & 67.2 & 30.1 & 26.8 & 34.5 & 40.2 & - & - \\
Open-TransMind \cite{shi2023open} & 512×512 & 83.4 & - & - & - & - & - & - & - \\
Transformer Fusion \cite{lou2023multi} & 640×640 & 85.4 & 73.6 & 39.5 & 43.6 & 43.7 & 58.6 & - & - \\
EMDFNet \cite{li2024emdfnet} & 640×640 & 84.4 & 76.8 & 64.9 & \textit{55.4} & 68.5 & 81.7 & 46.7 & 98.4 \\
YOLOv9 \cite{wang2024yolov9} & 640×640 & \textit{86.5} & - & - & - & - & - & 49.3 & 239.4 \\
YOLOv10 \cite{wang2024yolov10} & 640×640 & 85.2 & - & - & - & - & - & 66.5 & 127.6 \\
GOLD-YOLO \cite{goldyolo2023} & 640×640 & 84.9 & - & - & - & - & - & 43.0 & 151.7 \\
YOLOv8 \cite{jocher2023ultralytics} & 640×640 & \textbf{89.5} & - & - & - & - & - & 71.4 & 165.4 \\
YOLOv6 \cite{yolov6_meituan2023} & 640×640 & 84.4 & - & - & - & - & - & 52.2 & 150.7 \\
YOLOv5 \cite{jocher2020ultralytics} & 640×640 & 83.8 & - & - & - & - & - & \textit{82.6} & 109.1 \\
VATSD \cite{wang2023vatsd_tits} & 608×608 & 82.8 & - & - & - & - & - & \textbf{100.0} & \textbf{16.6} \\
I2D-Net \cite{fu2023tim_i2dnet} & 512×512 & 71.6 & - & - & - & - & - & - & - \\
Zhang et al.~\cite{zhang2024anchorfree_tetci} & 640×640 & 70.2 & - & - & - & - & - & - & 95.0 \\
\hline
\textbf{SMDDFNet (This paper)} & 640×640 & 86.4 & \textbf{79.5} & \textit{65.7} & \textbf{58.1} & \textbf{75.0} & \textbf{83.5} & 75.19 & \textit{81.9}\\

\hline
\end{tabular}
}
\label{tab:tt100k_results}
\end{table*}

\begin{table*}[!t]
\centering
\caption{Per-category AP on the TT100K test set (each column is one sign class). Baseline per-class results follow the protocol in~\cite{li2024emdfnet}; Faster R-CNN (ResNet) denotes the ResNet-backbone variant reported therein.}
\resizebox{\textwidth}{!}{
\begin{tabular}{lcccccccccccccc}
\hline
\textbf{Method} & \textbf{i2} & \textbf{i4} & \textbf{i5} & \textbf{il100} & \textbf{il60} & \textbf{il80} & \textbf{i2r} & \textbf{ip} & \textbf{p10} & \textbf{p11} & \textbf{p12} & \textbf{p19} & \textbf{p23} & \textbf{p26}\\
\hline
Faster R-CNN \cite{ren2015faster} & 44.0 & 46.0 & 45.0 & 41.0 & 57.0 & 62.0 & 41.0 & 39.0 & 45.0 & 38.0 & 60.0 & 59.0 & 65.0 & 50.0 \\
Faster R-CNN (ResNet) \cite{he2016deep} & 59.3 & 73.8 & 79.7 & 76.6 & 76.3 & 68.5 & 64.9 & 66.8 & 52.2 & 58.5 & 45.9 & 48.2 & 74.4 & 66.1 \\
FPN \cite{lin2017feature} & 72.5 & 79.6 & 88.3 & 90.2 & 88.2 & 84.9 & 77.4 & 75.8 & 62.7 & 75.9 & 60.2 & 53.7 & 75.8 & 76.0 \\
Mask R-CNN \cite{he2017mask} & 71.4 & 85.6 & 89.0 & 89.4 & 86.3 & 82.3 & 78.0 & 77.6 & 59.6 & 76.9 & 63.8 & 52.0 & 72.9 & 81.7 \\
SSD512 \cite{liu2016ssd} & 70.1 & 79.3 & 85.3 & 77.1 & 86.4 & 78.7 & 72.3 & 71.6 & 64.5 & 57.1 & 67.7 & 73.0 & 80.4 & 70.7 \\
DSSD512 \cite{fu2017dssd} & 65.0 & 86.2 & 88.6 & 62.7 & 87.7 & 76.2 & 60.2 & 85.5 & 66.2 & 55.1 & 54.4 & 78.4 & 79.3 & 75.5 \\
RFB Net 512 \cite{liu2018receptive} & 75.6 & 79.4 & 87.9 & 87.4 & 89.9 & 88.4 & 77.2 & 79.0 & 66.1 & 66.9 & 71.1 & 72.8 & 83.4 & 74.9 \\
ScratchDet \cite{zhu2019scratchdet} & 76.6 & 86.9 & 89.2 & 82.2 & 88.8 & 81.3 & 73.9 & 77.3 & 68.8 & 65.3 & 70.8 & 67.2 & 80.2 & 74.9 \\
CAB Net \cite{cui2020cabnet} & 76.0 & 87.5 & 89.4 & 80.6 & 89.9 & 85.3 & 80.5 & 78.0 & 69.1 & 77.6 & 74.3 & 87.6 & 87.1 & 81.4 \\
CAB-s Net \cite{cui2020cabnet} & 75.2 & 86.4 & 89.4 & 84.9 & 89.2 & 89.1 & 81.6 & 77.8 & 69.7 & 72.4 & 72.3 & \textbf{89.0} & 88.3 & 81.6 \\
\textbf{SMDDFNet (This paper)} & \textbf{88.3} & \textbf{92.0} & \textbf{93.9} & \textbf{95.1} & \textbf{98.3} & \textbf{95.7} & \textbf{91.8} & \textbf{94.5} & \textbf{72.1} & \textbf{87.6} & \textbf{79.7} & 80.1 & \textbf{90.7}  & \textbf{89.8}\\
\hline
\hline
\textbf{Method} & \textbf{p3} & \textbf{p5} & \textbf{p6} & \textbf{pg} & \textbf{ph4} & \textbf{ph4.5} & \textbf{pl100} & \textbf{pl120} & \textbf{pl20} & \textbf{pl30} & \textbf{pl40} & \textbf{pl5} & \textbf{pl50} & \textbf{pl60} \\
\hline
Faster R-CNN \cite{ren2015faster} & 48.0 & 57.0 & 75.0 & 80.0 & 68.0 & 58.0 & 68.0 & 67.0 & 51.0 & 43.0 & 52.0 & 53.0 & 39.0 & 53.0 \\
Faster R-CNN (ResNet) \cite{he2016deep} & 65.4 & 74.9 & 39.1 & 78.2 & 58.0 & 36.5 & 77.6 & 74.6 & 40.5 & 48.5 & 60.2 & 65.4 & 49.0 & 51.2 \\
FPN \cite{lin2017feature} & 71.6 & 79.2 & 39.1 & 78.2 & 58.0 & 36.5 & 87.5 & 85.5 & 55.7 & 55.6 & 71.5 & 77.3 & 60.8 & 58.7 \\
Mask R-CNN \cite{he2017mask} & 78.5 & 78.9 & 48.3 & 88.5 & 63.9 & 58.1 & 86.7 & 82.4 & 58.6 & 53.3 & 68.2 & 76.4 & 63.5 & 56.6 \\
SSD512 \cite{liu2016ssd} & 66.5 & 74.9 & 63.9 & 84.2 & 62.1 & 51.2 & 85.1 & 84.2 & 45.4 & 66.6 & 65.7 & 60.5 & 58.3 & 64.0 \\
DSSD512 \cite{fu2017dssd} & 56.1 & 79.6 & 55.4 & 85.8 & 60.7 & 88.6 & 79.1 & 69.6 & 65.3 & 68.3 & 68.2 & 61.5 & 65.5 & 64.7 \\
RFB Net 512 \cite{liu2018receptive} & 69.0 & 77.6 & 68.8 & 88.9 & 67.6 & 63.0 & 88.8 & 84.9 & 66.8 & 71.8 & 71.6 & 75.0 & 62.9 & 70.4 \\
ScratchDet \cite{zhu2019scratchdet} & 71.2 & 87.3 & 65.4 & 79.1 & 66.8 & 55.7 & 85.8 & 84.7 & 63.6 & 67.1 & 73.2 & 65.4 & 69.9 & 72.8 \\
CAB Net \cite{cui2020cabnet} & 74.7 & 84.5 & 82.5 & 87.5 & 71.8 & 64.4 & 88.4 & 87.9 & 68.6 & 73.3 & 74.8 & 79.3 & 75.1 & 76.3 \\
CAB-s Net \cite{cui2020cabnet} & 76.8 & 85.4 & 78.0 & 86.4 & 71.7 & 62.3 & 89.2 & 88.7 & 71.5 & 73.5 & 75.3 & 75.9 & 73.1 & 75.9 \\
\textbf{SMDDFNet (This paper)} & \textbf{83.2} & \textbf{89.3} & \textbf{83.0} & \textbf{97.3} & \textbf{73.2} & \textbf{90.3} & \textbf{93.9} & \textbf{93.6} & \textbf{79.4} & \textbf{86.6} & \textbf{85.1} & \textbf{86.3} & \textbf{81.4} & \textbf{84.6} \\
\hline
\hline
\textbf{Method} & \textbf{pl80} & \textbf{pm20} & \textbf{pm30} & \textbf{pm55} & \textbf{pn} & \textbf{pne} & \textbf{p13} & \textbf{pr40} & \textbf{w13} & \textbf{w55} & \textbf{w57} & \textbf{w59} & \textbf{p27} & \textbf{pl70} \\
\hline
Faster R-CNN \cite{ren2015faster} & 52.0 & 61.0 & 67.0 & 61.0 & 37.0 & 47.0 & 37.0 & 75.0 & 33.0 & 39.0 & 48.0 & 39.0 & 79.0 & 61.0 \\
Faster R-CNN (ResNet) \cite{he2016deep} & 59.0 & 50.5 & 29.1 & 68.5 & 77.8 & 87.5 & 47.7 & 86.9 & 30.9 & 62.1 & 67.0 & 57.2 & 64.3 & 61.2 \\
FPN \cite{lin2017feature} & 70.9 & 55.5 & 40.1 & 75.7 & 89.0 & 89.8 & 60.2 & 87.6 & 45.3 & 65.9 & 70.3 & 62.3 & 84.8 & 63.5 \\
Mask R-CNN \cite{he2017mask} & 71.5 & 58.0 & 41.5 & 68.8 & 88.6 & 90.5 & 63.0 & 87.5 & 51.3 & 66.6 & 71.1 & 61.8 & 87.5 & 66.3 \\
SSD512 \cite{liu2016ssd} & 70.5 & 69.6 & 51.3 & 71.2 & 71.7 & 86.4 & 51.8 & 87.9 & 46.1 & 64.6 & 74.0 & 58.8 & 76.2 & 70.6 \\
DSSD512 \cite{fu2017dssd} & 75.6 & 66.3 & 50.6 & 76.5 & 67.2 & 88.9 & 51.7 & 88.0 & 60.6 & 70.1 & 83.6 & 75.1 & 60.9 & 71.0 \\
RFB Net 512 \cite{liu2018receptive} & 71.9 & 73.7 & 54.0 & 86.5 & 78.0 & 88.2 & 59.8 & 84.5 & 64.8 & 72.4 & 81.5 & 69.3 & 79.8 & 64.9 \\
ScratchDet \cite{zhu2019scratchdet} & 75.9 & 73.3 & 52.2 & 76.5 & 76.7 & 89.4 & 62.9 & 85.0 & 69.1 & 70.3 & 84.7 & 76.5 & 79.7 & 70.2 \\
CAB Net \cite{cui2020cabnet} & 78.8 & 73.8 & 67.3 & 80.5 & 85.4 & 89.5 & 63.5 & 88.9 & 70.7 & 66.8 & 83.5 & 79.4 & 81.0 & 72.9 \\
CAB-s Net \cite{cui2020cabnet} & 78.4 & 71.2 & 67.0 & 83.3 & 82.2 & 89.2 & 64.0 & 88.2 & 57.2 & 75.2 & 80.1 & 66.6 & 87.5 & 70.7 \\
\textbf{SMDDFNet (This paper)} & \textbf{86.4} & \textbf{81.6} & \textbf{75.4} & \textbf{91.7} & \textbf{92.9} & \textbf{95.8} & \textbf{79.8} & \textbf{95.3} & \textbf{86.0} & \textbf{87.3} & \textbf{94.7} & \textbf{85.3} & \textbf{94.1} & \textbf{83.2} \\
\hline

\end{tabular}
}
\label{tab:tt100k_each_results}
\end{table*}

This work compares the proposed SMDDFNet with state-of-the-art object detection methods on the GTSDB dataset. Table~\ref{tab:gtsdb_results} presents the results in terms of mAP@.5, mAP@.5:.95, FPS, and GFLOPs. As shown, SMDDFNet achieves an mAP@.5 of 95.1\%, matching the best-performing R-FCN ResNet 101 (95.1\%) and slightly outperforming Faster R-CNN ResNet 101 (95.0\%) and ResNet 50 (91.5\%). Relative to the re-implemented YOLOv3 (93.8\%) and SSD (93.2\%) baselines summarized by Wu et al.~\cite{wu2022ssdrp}, as well as additional recent entries such as YOLO-PPA (93.2\%) and MASG-Net (90.8\%), SMDDFNet provides a clear margin on mAP@.5 while operating at higher FPS under our deployment settings. More importantly, SMDDFNet attains the highest mAP@.5:.95 of 79.3\%, clearly surpassing all other methods, including R-FCN ResNet 101 (75.2\%), Faster R-CNN ResNet 101 (74.1\%), and Cascade R-CNN + IFA-FPN (68.1\%), which highlights its robustness under stricter localization thresholds. In terms of inference speed, SMDDFNet achieves 75.19 FPS, far exceeding high-accuracy models such as Faster R-CNN ResNet 101 (8.1 FPS) and R-FCN ResNet 101 (11.7 FPS), while maintaining comparable or even superior detection accuracy. Although lightweight detectors like YOLO V2 (46.5 FPS) offer fast processing, they suffer from a considerable drop in accuracy (78.8\% mAP@.5, 60.0\% mAP@.5:.95); YOLO V2 also reports the lowest GFLOPs in the table, whereas SMDDFNet favors higher mAP with a moderate GFLOPs budget. Regarding efficiency, SMDDFNet incurs 81.9 GFLOPs, which is significantly lower than Faster R-CNN ResNet 50 (533.5 GFLOPs) and ResNet 101 (625.7 GFLOPs). This balance of high accuracy, real-time speed, and moderate complexity demonstrates that SMDDFNet provides an excellent trade-off between detection precision and computational efficiency, making it well-suited for real-world traffic sign detection applications.

\begin{table*}[!t]
\centering
\caption{Comparison with existing methods on the GTSDB dataset.}
\resizebox{\textwidth}{!}{
\begin{tabular}{lcccc}
\hline
\textbf{Model} & \textbf{mAP@.5} & \textbf{mAP@.5:.95} & \textbf{FPS} & \textbf{GFLOPs} \\
\hline
Faster R-CNN + IFA-FPN \cite{ren2015faster} & 78.0 & 65.4 & - & - \\
Faster R-CNN Inception V2 \cite{ren2015faster} & 90.6 & 70.2 & 17 & 120.6 \\
Faster R-CNN ResNet 50 \cite{ren2015faster} & 91.5 & 72.6 & 9.6 & 533.5 \\
Faster R-CNN ResNet 101 \cite{ren2015faster} & \textit{95.0} & 74.1 & 8.1 & 625.7 \\
R-FCN ResNet 101 \cite{dai2016r} & \textbf{95.1} & \textit{75.2} & 11.7 & 269.9 \\
YOLO V2 \cite{redmon2016you} & 78.8 & 60.0 & 46.5 & \textbf{62.7} \\
Cascade R-CNN \cite{cai2018cascade} & 70.4 & 52.7 & - & - \\
Cascade R-CNN + IFA-FPN \cite{cai2018cascade} & 80.3 & 68.1 & - & - \\
Improved YOLOv5 \cite{jocher2020ultralytics} & 81.9 & - & 30.4 & - \\
YOLO-PPA \cite{zhang2024yolo} & 93.2 & - & - & - \\
MASG-Net \cite{du2025lightweight} & 90.8 & - & - & - \\
YOLOv3 \cite{wu2022ssdrp} & 93.8 & - & 28.6 & - \\
SSD (VGG-16) \cite{wu2022ssdrp} & 93.2 & - & \textit{52.6} & - \\
\textbf{SMDDFNet (This paper)} & \textbf{95.1} & \textbf{79.3} & \textbf{75.19} & \textit{81.9} \\
\hline
\end{tabular}
}
\label{tab:gtsdb_results}
\end{table*}

This work further evaluates the proposed SMDDFNet on the VOC dataset, comparing it with state-of-the-art object detection models as shown in Table~\ref{tab:voc_results}. SMDDFNet achieves a mAP@.5 of 85.0\%, outperforming YOLOv7 (85.0\%), YOLOv5 (80.1\%), and YOLOX (81.0\%), demonstrating competitive performance among recent detectors. For stricter localization criteria, SMDDFNet attains a mAP@.5:.95 of 66.9\%, surpassing YOLOv7 (70.0\%) and YOLOv5 (65.2\%), indicating strong robustness across different IoU thresholds. It delivers the strongest listed scores on this benchmark. The Dynamic Dual Fusion (DDF) module in SMDDFNet, while designed to enhance detection in specialized tasks such as traffic sign recognition, still provides reliable generalization, making the model versatile across both specialized and general detection scenarios.

\begin{table}[!t]
\centering
\caption{Comparison with existing methods on the PASCAL VOC 2007 test set.}
\footnotesize
\setlength{\tabcolsep}{4pt}
\begin{tabular}{@{}lcc@{}}
\toprule
\textbf{Model} & \textbf{mAP@.5} & \textbf{mAP@.5:.95} \\
\midrule
Faster R-CNN \cite{ren2015faster} & 76.4 & 58.2 \\
YOLOv2 \cite{redmon2016you} & 73.0 & 57.0 \\
YOLOv4 \cite{bochkovskiy2020yolov4} & 77.8 & 61.5 \\
YOLOv5 \cite{jocher2020ultralytics} & 80.1 & 65.2 \\
YOLOX \cite{yolox2021} & \textit{81.0} & 47.2 \\
YOLOv7 \cite{wang2023yolov7} & \textbf{85.0} & \textbf{70.0} \\
\textbf{SMDDFNet (This paper)} & \textbf{85.0} & \textit{66.9} \\
\bottomrule
\end{tabular}
\label{tab:voc_results}
\end{table}

To further verify the detection performance of the proposed SMDDFNet in vehicle detection scenarios, we conduct comparative experiments on the vehicle dataset, with detailed quantitative results presented in Table~\ref{tab:vehicle_results}. In terms of mAP@0.5, SMDDFNet reaches 47.62\%, outperforming YOLOv8 (46.95\%), YOLOv7 (45.60\%), YOLOX (45.12\%), YOLOv5 (45.40\%), and YOLOv4 (43.28\%), achieving the highest overall detection accuracy among the listed detectors. In terms of mAP@0.5:0.95 that evaluates comprehensive localization precision, our model obtains 34.41\%, which is superior to YOLOv8 (33.55\%), YOLOv7 (33.25\%), and the remaining baselines, demonstrating strong multi-scale positioning capability under complex IoU constraints. For precision, recall, and F1, SMDDFNet achieves 51.19\%, 59.35\%, and 54.97\%, respectively: recall is the highest in the table, whereas precision is second to YOLOv7 (55.50\%) and F1 is slightly below YOLOv8 (55.25\%), reflecting a recall-oriented operating point alongside the best mAP scores. These results corroborate the effectiveness of the proposed design on an in-the-wild vehicle benchmark.

\begin{table*}[!t]
\centering
\caption{Comparison with existing methods on the vehicle validation set. }
\resizebox{\textwidth}{!}{
\begin{tabular}{lccccc}
\hline
\textbf{Model} & \textbf{mAP@0.5} & \textbf{mAP@0.5:0.95} & \textbf{P (\%)} & \textbf{R (\%)} & \textbf{F1 (\%)} \\
\hline
YOLOv4 \cite{bochkovskiy2020yolov4} & 43.28 & 30.82 & - & - & - \\
YOLOv5 \cite{jocher2020ultralytics}    & 45.40 & 32.15 & 50.26 & 52.73 & 51.42 \\
YOLOX \cite{yolox2021}            & 45.12 & 32.68 & - & - & - \\
YOLOv7 \cite{wang2023yolov7}     & 45.60 & 33.25 & \textbf{55.50} & 53.90 & 54.69 \\
YOLOv8 \cite{jocher2023ultralytics}  & \textit{46.95} & \textit{33.55} & \textit{52.13} & \textit{58.84} & \textbf{55.25} \\
\textbf{SMDDFNet (This paper)} & \textbf{47.62} & \textbf{34.41} & 51.19 & \textbf{59.35} & \textit{54.97} \\
\hline
\end{tabular}
}
\label{tab:vehicle_results}
\end{table*}

\subsection{Qualitative analysis}
\begin{figure}[!t]
    \centering
    \includegraphics[width=0.75\linewidth]{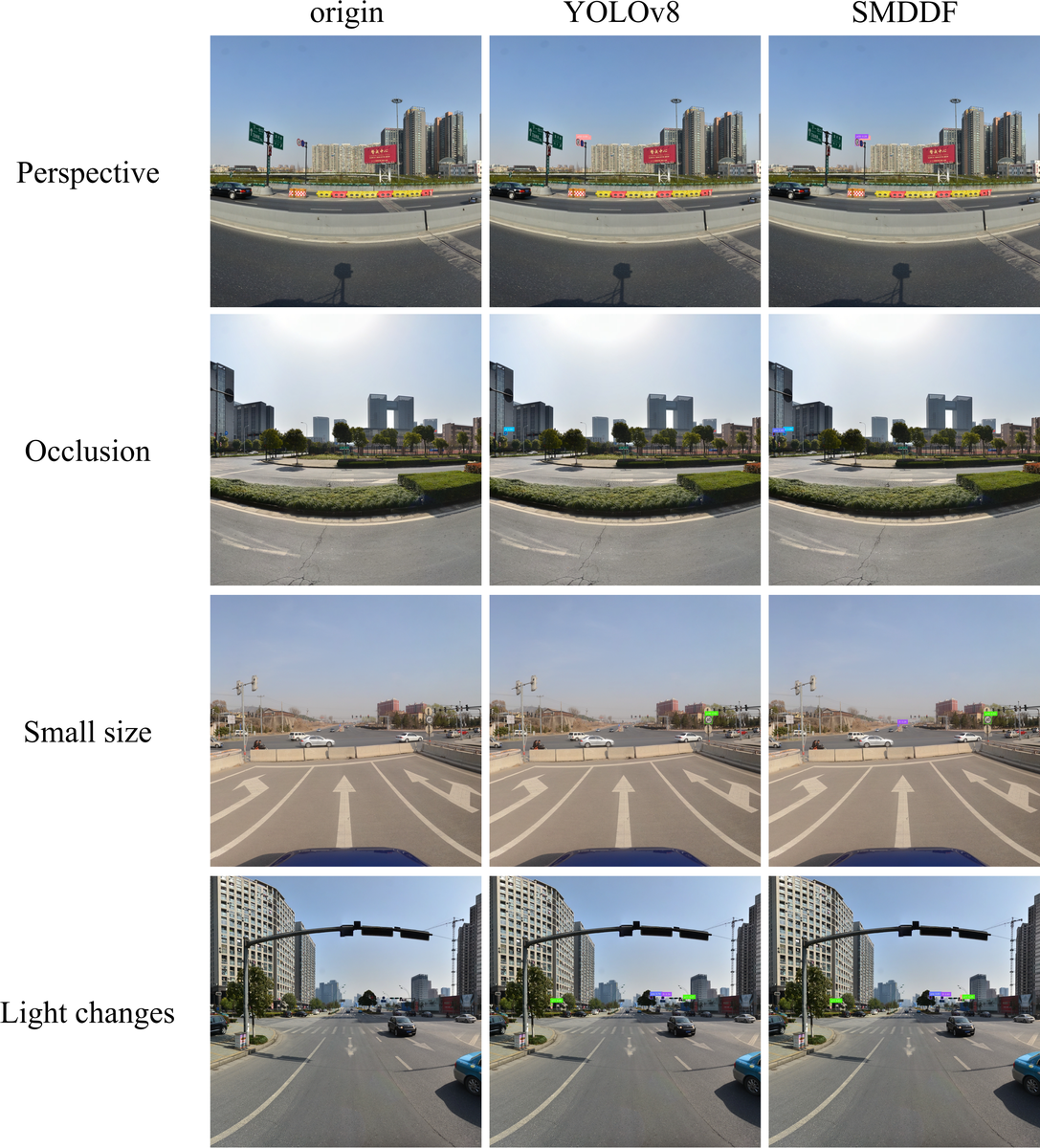}
    \caption{Comparison of detection performance between SMDDFNet and baseline on the TT100K testing set.}
    \label{fig:qualitative_TT100K}
\end{figure}

To further validate the effectiveness of the proposed SMDDFNet, we conduct qualitative comparisons between SMDDFNet and the baseline YOLOv8 on challenging traffic sign scenarios. As illustrated in Figure~\ref{fig:qualitative_TT100K}, SMDDFNet demonstrates superior detection performance in complex conditions, including small-scale targets, varying illumination, occlusion, and viewpoint changes. Specifically, in the second column of Figure~\ref{fig:qualitative_TT100K}, SMDDFNet accurately detects small traffic signs that are missed or incorrectly classified by the baseline, highlighting its enhanced capability in small object perception. Additionally, under low-light conditions or overexposure, this model maintains stable detection results due to the effective multi-scale representation and dynamic fusion introduced by the DDF module. In the third column of Figure~\ref{fig:qualitative_TT100K}, SMDDFNet also shows stronger robustness against partial occlusion and large-angle view changes, where baseline predictions are often inaccurate or incomplete. These qualitative results further support the quantitative improvements observed in previous sections, demonstrating that SMDDFNet is more adaptable and reliable for real-world traffic sign detection tasks.

\begin{figure}[!t]
    \centering
    \includegraphics[width=0.75\linewidth]{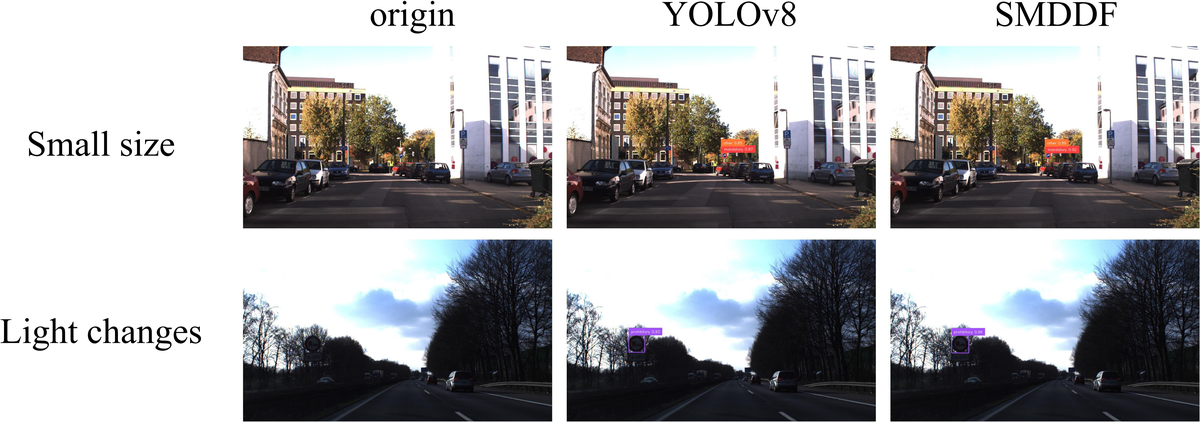}
    \caption{Comparison of detection performance between SMDDFNet and baseline on the GTSDB testing set.}
    \label{fig:qualitative_GTSDB}
\end{figure}

\begin{figure}[!t]
    \centering
    \includegraphics[width=0.75\linewidth]{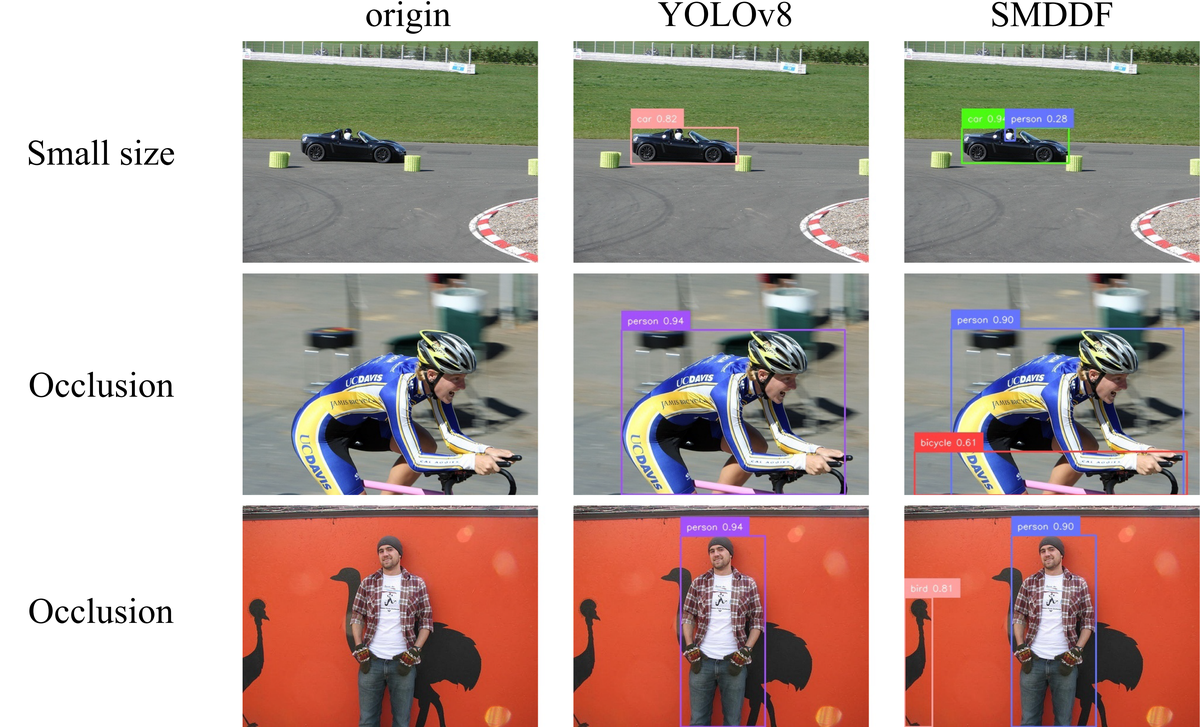}
    \caption{Comparison of detection performance between SMDDFNet and baseline on the PASCAL VOC 2007 test set.}
    \label{fig:qualitative_VOC}
\end{figure}

\begin{figure}[!t]
    \centering
    \includegraphics[width=0.75\linewidth]{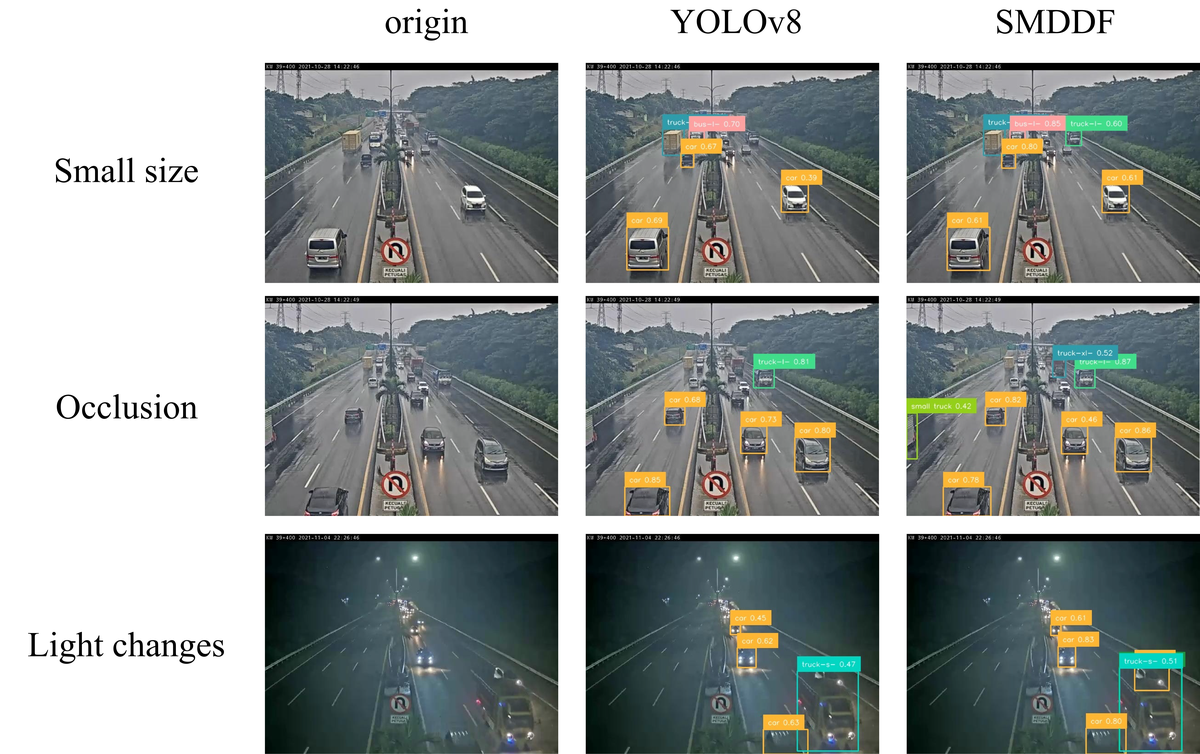}
    \caption{Comparison of detection performance between SMDDFNet and baseline on the vehicle test set.}
    \label{fig:qualitative_vehicle}
\end{figure}

\subsection{Ablation analysis}
To comprehensively evaluate the contribution of each component within SMDDFNet, this study conducts a series of ablation experiments. Specifically, Efficient Multi-scale Attention (EMA), Dynamic Filter (DF), the combined Dynamic Dual Fusion (DDF), and the State-space Modeling backbone with the Multi-Scale Feature Fusion (MSFF) neck are progressively integrated into the YOLOv8 baseline. By adding each module stepwise, we isolate and quantify its impact on detection performance. All experiments use consistent training settings on the TT100K dataset to ensure comparability; results are summarized in Table~\ref{tab:ablation_tt100k}. 

\begin{table}[!t]
  \centering
  \caption{Ablation study on TT100K for EMA, DF, DDF, and the state-space backbone with MSFF neck.}
  \resizebox{0.8\textwidth}{!}{
  \begin{tabular}{lcccccc}
    \hline
    Model & mAP@.5 & mAP@.75 & mAP@.5:.95 & $AP_s$ & $AP_m$ & $AP_l$ \\
    \hline
    Baseline & 81.3 & 73.2 & 61.8 & 53.6 & 66.5 & 73.0 \\
    Baseline+EMA & 82.5 & 74.5 & 62.9 & 54.2 & 67.9 & 74.2 \\
    Baseline+DF & 83.2 & 75.1 & 63.7 & 55.8 & 68.7 & 76.1 \\
    Baseline+DDF & 83.9 & 75.8 & 64.5 & 56.5 & 69.4 & 77.3 \\
    Baseline+backbone+neck & 85.4 & 77.0 & \textbf{65.9} & 57.3 & 71.0 & 78.8 \\
    \hline
    SMDDFNet (This paper) & \textbf{86.4} & \textbf{79.5} & 65.7 & \textbf{58.1} & \textbf{75.0} & \textbf{83.5} \\
    \hline
  \end{tabular}
}
  \label{tab:ablation_tt100k}
\end{table}

\textbf{Impact of efficient multi-scale attention (EMA).} To assess how the EMA branch improves spatial aggregation, EMA is inserted into the baseline. As shown in Table \ref{tab:ablation_tt100k}, mAP@.5 increases by 1.2\%, mAP@.75 increases by 1.3\%, and mAP@.5:.95 increases by 1.1\%. However, while overall detection performance improves, the relatively small increase in AP$_s$ (+0.6\%) suggests that EMA has a limited impact on the detection of small objects.

\textbf{Impact of dynamic filter (DF).} To improve adaptability to varying object shapes and scales, DF is incorporated into the baseline. Compared to the baseline, DF enhances mAP@.5 by 1.9\%, mAP@.75 by 1.9\%, and mAP@.5:.95 by 1.9\%. In particular, AP$_s$ increases by 2.2\%, demonstrating the effectiveness of DF on small objects, while improvements in AP$_m$ (+2.2\%) and AP$_l$ (+3.1\%) confirm benefits at medium and large scales as well.

\textbf{Impact of dynamic dual fusion.} Expanding upon DF, the DDF module introduces a hierarchical fusion strategy to integrate multi-scale contextual features dynamically. The experimental results indicate that DDF further improves mAP@.5 by 0.7\%, mAP@.75 by 0.7\%, and mAP@.5:.95 by 0.8\% over DF. Furthermore, AP$_s$, AP$_m$, and AP$_l$ increase by 0.7\%, 0.7\%, and 1.2\%, respectively, highlighting the gains from jointly fusing EMA and DF in DDF.

\textbf{Impact of backbone and neck.} To enhance long-range feature representation, this study replaces the standard YOLOv8 backbone with the State-space Modeling Network and adopts the MSFF neck for multi-scale fusion. Compared to the previous model, this backbone and neck achieve a significant boost of 1.5\% in mAP@.5, 1.2\% in mAP@.75, and 1.4\% in mAP@.5:.95. Improvements in AP$_m$ (+1.6\%) and AP$_l$ (+1.5\%) confirm its ability to capture long-range dependencies and enhance feature extraction for medium and large objects.

\section{Conclusion}
This paper proposes SMDDFNet, a traffic sign detection framework for visual signal processing in complex driving scenes, targeting small objects, multi-scale representation, and real-time inference. A lightweight State-space Modeling backbone captures local and global context efficiently; the Dynamic Dual Fusion (DDF) module combines Efficient Multi-scale Attention with dynamic frequency-domain filtering to adapt features across scales; and the Multi-Scale Feature Fusion (MSFF) module aggregates pyramid features for robust localization. On TT100K, SMDDFNet achieves strong accuracy for small and occluded signs together with practical real-time throughput relative to accuracy-oriented baselines. Evaluations on GTSDB, PASCAL VOC, and the Roboflow~100 \emph{vehicle} subset further indicate generalization beyond curated traffic-sign splits. Future work may include tighter integration with embedded platforms and domain adaptation across regions and sensor modalities.

\section*{CRediT authorship contribution statement}
\textbf{Tianyi Yu:} Conceptualization, Methodology, Software, Writing – original draft, Writing – review \& editing.
\textbf{Dajian Zhong:} Supervision, Funding acquisition, Writing – review \& editing, Project administration, Validation.
\textbf{Lilin Wang:} Supervision, Writing – review \& editing.

\section*{Declaration of competing interest}
The authors declare that they have no known competing financial interests or personal relationships that could have appeared to influence the work reported in this paper.

\section*{Funding}
This work was supported by the Shanghai Sailing Program [grant number 24YF2716400].

\section*{Data availability}
TT100K is available at \url{https://cg.cs.tsinghua.edu.cn/traffic-sign/}. GTSDB is available at \url{https://benchmark.ini.rub.de/gtsdb_dataset.html}. PASCAL VOC can be accessed from \url{http://host.robots.ox.ac.uk/pascal/VOC/}. The \emph{vehicle} split is part of the Roboflow~100 benchmark.

\section*{Declaration of generative AI and AI-assisted technologies in the writing process}
During the preparation of this work, the authors did not use generative AI or AI-assisted technologies to create scientific content, experimental results, figures, or conclusions. Basic grammar and spelling checks were applied under full author review and responsibility.

\bibliographystyle{elsarticle-num}
\bibliography{bib}

\end{document}